\documentclass{article}
\usepackage[utf8]{inputenc}
\usepackage[nottoc]{tocbibind}
\usepackage[english]{babel}
\usepackage{natbib}
\usepackage{color}
\usepackage{hyperref}
\usepackage{url}
\usepackage{enumitem}
\usepackage{amsmath}
\usepackage{amssymb}
\usepackage{geometry}
\usepackage{setspace} 
\newcommand{\note}[1]{#1}

\title{Survey on safe robot control via learning}
\author{Bassel El Mabsout}
\date{September 2022}

\begin{document}

\maketitle

\section{Introduction}
    Modern society heavily relies on robotic systems, their use affects the aerospace, automotive, energy, disaster response, health care, manufacturing, and traffic management industries among countless others.
    From making robots walk \cite{bipedal_control_book} to getting molecular swarms to kill cancer cells \cite{cell_control}, whole fields of research dedicate themselves to the problem of control.
    Intelligently selecting control strategies so that we can manage, direct, or command the trajectories a system can take distills the essence of problems faced in control.
    When a system can be controlled in the aforementioned manner using control loops, the system in question is termed a \emph{control system}.
    Tackling the problem of control, the research community has produced many alternative solutions with varying trade-offs concerning what is achievable and how much we can represent these systems and our goals. Thus, research works that attack control problems sit on a wide spectrum of solutions.
    On one end, \textbf{classical control theory} commonly defines these systems by modeling their evolution using differential equations as a function of their free parameters.
    Such techniques involve working with a subset of mathematically definable models and objective functions.
    The intention is to make the selection of these control parameters representable as a problem with efficiently computable solutions.
    On the other side of the spectrum, \textbf{Deep Reinforcement Learning} (Deep-RL) algorithms define and incrementally improve "policies" (or analogously "controllers") on set objectives defined by loss functions via gradient updates.
    Such solutions may be entirely data-driven, meaning that controllers can improve without any prior model defined, thereby only requiring experience gathered from real or modeled systems via observations. 
    Of course, there are also large bodies of work exploring the design space between these two extremes.

    When goals include safety concerns, special care must be taken to prevent the system from ever reaching states which violate them.
    In this document, we focus on the problem of learning high-performance control on real robots while maintaining various notions of safety.
    Such notions include stability, obstacle avoidance, over-actuation prevention, and liveliness properties. In real systems, these notions of safety map to the ability to prevent ovens from burning down the house or preventing collisions in air traffic control systems.
    Section~\ref{sec:classical_control_theory} describes and categorizes classical control theory techniques, presenting their challenges and guarantees.
    Section~\ref{sec:learning_based_control} presents learning-based methods for control, focusing on RL and augmentations of classical control theory.
    Section~\ref{sec:embedded_design} then discusses some embedded system hardware and software considerations for retaining safety within complex robotic systems.


\section{\note{Safety and Control in Classical Control Theory}}\label{sec:classical_control_theory}
    There is a wealth of classical control techniques, commonly categorized as prescriptive/descriptive, model-free/model-based, myopic/predictive, and discrete/continuous \cite{Franklin2001-xe, Dorf2010-cx}.
    Descriptive methods analyze whether given controllers follow defined specifications, while Prescriptive ones synthesize controllers with specified properties.
    Essential to these methods is whether we have access to predictive models representing the dynamics of the systems in question.
    We focus on the differences between myopic, predictive, model-free, and model-based control in Section~\ref{sub:ct_model_free_vs_model_based}. And on how such systems are modeled in Section~\ref{sub:ct_modeling}.

    \subsection{Model-free vs Model-based}\label{sub:ct_model_free_vs_model_based}
        Model-free controllers are defined as controllers with no access to dynamical models when making control decisions.
        They make weak assumptions concerning the control responses of the systems in question.
        For example, increasing specific outputs leads to a general increase in some variables in the system.
        Thus, even when the systems in question are poorly characterized, they offer a practical solution as controllers.
        Of such controllers, the \textbf{Proportional Integrative Derivative} (PID) \cite{pid} class of controllers is by far the most widely studied and used.
        Without access to models, such controllers turn to heavy use of feedback and often require careful tuning for stable and performant control.
        These controllers are considered myopic, meaning they must optimize the system's trajectory given only limited information prior to the current execution loop.
        Many of these systems are tuned by expert practitioners using different methods \cite{pid_tuning}---for PID, the most common is the Ziegler–Nichols method \cite{ziegler_nichols}.
        In this space, complex live-automated tuning methods \cite{adaptive_pid} also exist, allowing these systems to be adaptive to environmental changes.
        These methods tend to have low computational requirements for executing their control loop; trajectory planning algorithms (high-level controllers) commonly use them for reaching low-level objectives.
        Without a model, these controllers can fail to take advantage of system-specific behaviors, leading to under-performance and even complete failure in controlling some systems \cite{pid_limitations}.
        
        Verification is a common requirement in safety-critical systems, such as a car's brake control system.
        Verification necessitates the definition of a model accurate enough so that guarantees can be made about the controller reaching the specified objectives.
        For example, under the dynamics of an inverted pendulum with the goal of setting it upright, \note{\textit{The existence}} of a control Lyapunov function \cite{Isidori_1995} guarantees that a PID controller is stable in that it always reaches this goal from any starting angle.
        Prescriptive methods use a model to synthesize \emph{correct-by-construction} controllers built to satisfy the defined specification.
        Classically, assumptions on the mathematical structure of the models and the specifications made allow efficient optimization procedures for finding controllers.
        For example, under the assumption that a certain system's dynamics forms linear differential equations, and that the intended goal is to minimize a cost represented by a quadratic function, the theoretically accepted optimal solution is a feedback controller known as the \textbf{linear–quadratic} regulator (LQR) \cite{lqr}.
        When specifications include constraints that must be solved online, \textbf{Model Predictive Controllers} (MPCs)\cite{mpc} are used to iteratively solve a constrained optimization problem over finite time horizons.
        There is a trade-off between the computational complexity of such methods, assumptions made over defined models, and the expressive power of specification.
        Such methods also require system identification techniques to minimize errors in prediction.
        These controllers can be made adaptive by performing system identification online.
        State-of-the-art complex controllers use a mix of discrete and continuous models termed hybrid automata \cite{hybrid_automata}, complex specifications such as \textbf{Linear Temporal Logic} (LTL) \cite{ltl}, and geometric techniques for proving reachability properties controllers using, for example, Zonotopes \cite{zonotopes}.
        A control theoretic subfield exists for adaptive controllers, for example, the theory of L1 adaptive controllers \cite{l1_adaptive}.
    \subsection{Modeling and System Identification}\label{sub:ct_modeling}
        In continuous-time systems, dynamical models are commonly defined as ordinary or partial differential equations with free parameters that identify real-life systems.
        These systems can be time-invariant, linear, affine, stochastic, ergodic, and more.
        Many techniques have been developed to adapt well-understood techniques to a larger set of systems, for example, linearizing nonlinear systems around equilibrium points \cite{Isidori_1995}.
        Even when models represent well the primary components of a real-life system's time evolution, some parameters require identification, such as the weight and length of a pendulum.
        
        System identification methods have been well-explored theoretically, especially for linear time-invariant systems \cite{control_book, sys_id}.
        Data-driven methods use tuning procedures based on observations gathered on real-life systems and produce models with the best fit.
        This procedure "identifies" which system is operating in the real world.
        Examples of well-established techniques in this area include optimizing models with respect to assumptions of adversarial noise~\cite{adversarial_noise} and subspace methods~\cite{subspace} for handling systems with multiple inputs and outputs.
        These procedures are often executed offline, and then models are deployed on real-life systems for controller synthesis.
        In allowing for environmental changes, other approaches \cite{sys_id} allow for the system identification procedures to be executed online, leading to controllers adaptive to environmental changes.
        
        There are many complex systems where defining such models remains an open problem \cite{open_probs}, and even when possible, the evaluation of such models can often be computationally impractical \cite{navier_stokes}.
 
    \subsection{Example system}\label{sub:oven}
        
        A simple example of a control system would be an oven with a temperature sensor and a heater which can be turned on or off.
        According to Newton's law of cooling, the temperature in this system follows this equation:
        $$ \dot{temp}(t) = -k(temp(t) - temp_\text{ambient})$$
        $\dot{temp}(t)$ defines the change in temperature of the system at time $t$, $temp_\text{ambient}$ is the ambient temperature of the oven, this is the temperature that the oven reaches eventually.
        the constant $k \in \mathbb{R}^+$ is the rate at which the temperature of the oven closes the temperature "gap".
        In solving this differential equation, we first solve for the n'th derivative of $temp(t)$, namely $\overset{(n)}{temp}(t)$:
        \begin{align*}
            \dot{temp}(t) &= -k(temp(t) - temp_\text{ambient})\\
            \ddot{temp}(t) &= \dot{temp}(t)\frac{d}{dt} = -k(temp(t) - temp_\text{ambient})\frac{d}{dt} = -k*\dot{temp}(t) = (-k)^2(temp(t) - temp_\text{ambient})\\
            \dddot{temp}(t) &= \ddot{temp}(t)\frac{d}{dt} = (-k)^2(temp(t) - temp_\text{ambient})\frac{d}{dt} = (-k)^2\dot{temp}(t) = (-k)^3(temp(t) - temp_\text{ambient})\\
            \vdots\\
            \overset{(n)}{temp}(t) &=\overset{(n-1)}{temp}(t)\frac{d}{dt} = (-k)^{n-1}(temp(t) - temp_\text{ambient})\frac{d}{dt} =  (-k)^n(temp(t) - temp_\text{ambient})
        \end{align*}
        We now use define $temp(t)$ by its Maclaurin series (Taylor series expanded around $0$), and plug in the previous definition for the n'th derivatives.
        \begin{align*}
            temp(t) &= \sum^\infty_{i=0} \frac{\overset{(i)}{temp(0)}t^i}{i!} = temp(0) + \sum^\infty_{i=1} \frac{(-k)^i(temp(0) - temp_\text{ambient})t^i}{i!}\\
                    &= temp(0) + (temp(0) - temp_\text{ambient})\sum^\infty_{i=1} \frac{(-kt)^i}{i!}\\
                    &= temp(0) - (temp(0) - temp_\text{ambient}) + (temp(0) - temp_\text{ambient})\sum^\infty_{i=0} \frac{(-kt)^i}{i!}\\
                    &= temp_\text{ambient} + (temp(0) - temp_\text{ambient})e^{-kt}\hspace{1.8cm} (\text{since } e^x = \sum^\infty_{i=0} \frac{x^i}{i!})
        \end{align*}
\color{black}
        This follows the form of a \textbf{globally exponentially stabilizing system~[def.a]}.
        In trying to control this oven, we introduce an electric heating component which controls its ambient temperature, so now $temp_\text{ambient} = u(temp_\text{on} - temp_\text{off}) + temp_\text{off}$.
        The $u\in [0,1]$ parameter controls how much electricity we put in to heat up the oven.
        When $u$ is 0, $temp_\text{ambient}=temp_\text{off}$ which represents the temperature that the oven would reach if the heater is turned off, and when $u$ is 1.
        $temp_\text{ambient}=temp_\text{on}$, represents the temperature that the oven would reach when the heater is maximally on.
        Assuming we have precise control over the amount of electricity put into the heater, $u$ can take values between 0 and 1.
        We want the oven to reach a certain temperature $temp_\text{desired}$, and we can choose controllers $c$ of the form $u = c(temp(t), t)$, setting the value of $u$ at time $t$ according to an observed temperature $temp(t)$.
        We choose to analyze the following example controller: $c(temp(t), t) = u = \frac{temp_\text{desired} - temp_\text{off}}{temp_\text{on} - temp_\text{off}}$.
        Plugging in $u$, our transformed differential equation becomes:
        $$ \dot{temp}(t) = -k*(temp(t) - temp_\text{desired})$$
    
    \subsection{\note{Safety in Classical Control}}
        The meaning of \textit{safe} control depends on the type of \textbf{specification} used in representing safety and the specific dynamical system in question. In general, the goal is to constrain the system's evolution, preventing the violation of some safety criteria.
        For example, if we have an oven temperature controller, safety can mean never reaching temperatures higher than some amount (in preventing damage to some internal components).
        This safety criterion can be defined, for example, as an inequality separating the state space into a safe and unsafe set ($\forall_t, temp(t) < 400F$).
        A safe controller, in this case, never allows the system to reach points in the unsafe set.
        This is how the notion of \textit{reachability} connects to system safety.
        Alternatively, it can mean proving that the oven's temperature eventually converges to the one we desire ($temp(\infty) = temp_\text{desired} = 200F$).
        This convergence property would be a notion of stability.
        Unsafety here is defined as $\exists_t, ||temp(t+\epsilon) - temp_\text{desired}|| > ||temp(t) - temp_\text{desired}||$, meaning the temperature diverges from our desired one.
        Another example would be a Segway-like robot.
        An unstable controller would cause the error in the Segway reaching our desired objective (of staying upright) to increase over time, sometimes causing failures such as the Segway falling and causing harm to the rider, exemplifying how stability is often a safety criterion.
        \subsubsection{Stability}
            \paragraph{Definitions}
                \begin{enumerate}[label={[def.\alph*]},leftmargin=1.5cm]
                    \item \textbf{Global exponential stability} denotes the notion that a differential equation starting from any point $x(t_0)$ in the initial state space $\mathcal{X}$ will eventually ($t \to \infty$) reach the point $x(\infty)=0$ and that the rate of convergence is exponential, namely: $$\forall_{t\in \mathbb{R}^+, x(t_0) \in \mathcal{X}}, \exists_{m, \alpha \in \mathbb{R}^+},||x(t)|| \leq me^{\note{-}\alpha(t-t_0)}||x(t_0)||$$
                    \item \textbf{Control Lyapunov Functions} are functions $V$ which satisfying the following properties:
                    \begin{align}
                        V(p) &= 0\\
                        \forall_{x\in\mathcal{X/} \{p\}}V(x) & > 0\\
                        \forall_{x\in\mathcal{X}}\exists_{u\in\mathcal{U}}, V(f(x,u)) &\leq V(x)
                    \end{align}
                    Specifically, these are the conditions for a discrete-time Lyapunov control function.
                    $f$ represents the system's transition function, meaning $x'=f(x,u)$ defines the next state when starting from state $x$ and taking action $u$.
                    $\mathcal{X}$ defines the state-space, and $\mathcal{U}$ defines the action space.
                    Intuitively, $V$ is positive everywhere except the point we want the system to reach, which is $p$ (eq. (1) and (2)).
                    And there is always an action $u$ we can take that forces $V$'s value to be non-increasing when a system transition is taken (eq. (3)).
                    This means that with an infinite number of transitions we can always cause the system to eventually reach the setpoint $p$.
                    We can make the Lyapunov function be a proof of \textbf{global exponential stability} by changing eq. (3) such that $V(f(x,u)) - V(x) < \alpha V(x)$.

                    For the oven case with our controller $c$, we choose the following Lyapunov function:
                    $$V(temp(t)) = |temp(t) - temp_\text{desired}|$$
                    Proving the conditions:
                    \begin{align*}
                        &V(temp_\text{desired}) = |temp_\text{desired} - temp_\text{desired}| = 0\\
                        &\forall_{x\in\mathbb{R/} \{temp_\text{desired}\}}V(x) = |x - temp_\text{desired}| > 0 \text{ (since } x \neq temp_\text{desired})\\
                        &\forall_{x\in\mathbb{R}}, V(xe^{-k(\Delta t)}+temp_\text{desired}) \leq V(x) \leftrightarrow |xe^{-k\Delta t}| \leq |x| \leftrightarrow -k\Delta t < ln(1) \hspace{0.2cm}(k, \Delta t > 0\text{ as given})
                    \end{align*}

                    The oven system with controller $c$ is therefore a stable system with respect to reaching our desired temperature.
                \end{enumerate}
        \subsubsection{Temporal Logics}
            Formal logics such as temporal logics (ex. LTL, STL, rSTL, dSTL) define languages allowing for specifying behaviors we want the system to meet.
            Example of a Signal Temporal Logic (STL) formula for ovens:
            \begin{align*}
                \phi_\text{converge}(x,t) &= F(G(|x(t) - temp_\text{desired}| < 10)) \\
                \phi_\text{avoid}(x,t) &= G(x(t) > 400 \to F_{[0, 1s]}(G_{[0,10s]}(x(t) < 400))) \\
                \phi(x, t) &= \phi_\text{converge}(x,t) \wedge \phi_\text{avoid}(x,t)
            \end{align*}
            
            $G$ means \textit{globally} and $F$ means \textit{eventually}, $F(G(\alpha))$ represents the notion that after some undetermined amount of time, the statement $\alpha$ will be true and that it will stay true, in other words, eventually, $\alpha$ will be globally true.
            \note{When brackets are included, the operator is quantified over the time period specified, $F_{[0,1s]}\alpha$ means that the formula $\alpha$ will be "eventually" satisfied within a 1 second period.
            This notation $X[t_1,t_2]$ means that when the operator $X$ gets "activated" at time $t$, the operator applies from time $t+t_1$ to $t+t_2$.
            }
            Therefore the formula $\phi_\text{converge}$ denotes the behavior that the system eventually reaches and remains near the desired temperature (within 10 degrees Fahrenheit).
            $\phi_\text{avoid}$ represents the notion that, whenever the temperature reaches above 400F, within 1 second, the temperature will be dropped to under 400F, and will remain there for at least 10 seconds.
            This criteria bounds how long the oven spends at temperatures which we deem dangerous for the components.

            A Commonly used extension of Signal Temporal Logic adds robustness (rSTL), which allows for evaluating how far we are from satisfying the formula itself (ex. $200 - |x|$ represents a robustness value for the logical relation $|x| < 200$).
            This robustness value is often used to find controllers which are optimized to remain far from violating the STL specification.
            differentiable Signal Temporal Logic (dSTL) makes the robustness value differentiable so that gradient methods can be used for finding satisfying controllers (more prominently used for learning deep-NN controllers).


\section{Learning-based Control}\label{sec:learning_based_control}
    In addressing the limitations of classical control, learning-based methods have reached new milestones in solving complex control tasks. This includes grasping robotics \cite{rubik}, solving Atari games \cite{dqn}, and high-performance drone control \cite{real}.
    In Section~\ref{sub:deep_rl}, we focus on seminal works in the area of Deep-RL, which use benchmark example systems to test their methods' viability.
    Moreover, in Section~\ref{sub:augmented_ct}, we focus on recent works augmenting classical control theory techniques with machine learning methods.
    \subsection{Deep-RL}\label{sub:deep_rl}
        The central assumption in RL is that the system in question can be represented as a discrete-time \textbf{Markov Decision Process} (MDP) and that a reward function is defined to compensate policies for desirable behaviors.
        Such MDPs may have a continuous or discrete state-space and action-space.
        They may also be stochastic or deterministic. 
        Though policy search techniques such as Monte Carlo search-based methods \cite{alpha_go} rely on having explicit models,
        \textbf{Time Differencing} (TD) or \textbf{policy gradient} based methods \cite{Sutton_Barto_2018}, relieve this requirement, weakening the need for expert knowledge.
        Deep-RL, which uses deep learning for finding policies, represents controllers as deep NNs.
        These methods are often categorized into Model-"free" methods Section~\ref{sub:model_free_rl} or Model-based methods Section~\ref{sub:model_based_rl}.
        
        \subsubsection{Model-"free" RL}\label{sub:model_free_rl}
            Model-"free" RL algorithms do not model the MDP under consideration.
            However, a model is still involved in representing a policy's total expected discounted reward as a function of the system state.
            These algorithms are commonly categorized as Q-learning or Policy gradient-based, and whether they are On-policy or Off-policy algorithms.
            Modern Q-learning \cite{Qlearning} methods such as DQN\cite{dqn}, DDPG\cite{DDPG}, SAC\cite{SAC}, and TD3\cite{TD3}, rely on a learnt Q-value function otherwise known as a critic.
            The goal of critics is to represent the optimal policy's discounted expected reward given certain states and actions.
            Controllers are optimized to choose actions maximizing this Q-value. These methods are often off-policy, meaning they have a replay buffer for storing observed transitions in the environment, sampling batches from which to train the represented Q-value and policy.
            Being off-policy allows for having different policies used to explore an environment.
            On the other hand, policy gradient methods such as PPO\cite{PPO} and TRPO\cite{TRPO}, and A3C\cite{a3c} improve a stochastic controller's expected discounted reward by directly taking gradient steps on the policy's returns.
            These methods are often on-policy, performing gradient updates as new observations are gathered.
            One limitation of model-"free" RL works, especially policy-gradient-based methods, is that they are famously data-inefficient at producing performant policies.
    
        \subsubsection{Model-based RL}\label{sub:model_based_rl}
            A model allows agents to plan ahead, allowing the exploration of future paths with their respective returns in rewards.
            Like MPC in classical control theory, agents can make immediate choices in picking actions based on the chosen path representing the learned policy.
            AlphaGo \cite{alpha_go} is a famous example of this approach. When models are accurate, considerable improvements are observed in the sample efficiency for finding good policies.
            Other methods make use of a learned model \cite{model_based_rl} of the environment from observation data rather than assuming a predefined model.
            However, model-"free" RL still produces state-of-the-art policy performance in most accepted RL benchmarks \cite{rl_survey}.
    
        \subsubsection{Robustness in Deep-RL}
            In regular RL, the optimization goal is to find policies that maximize the discounted sum of rewards designed by expert practitioners.
            However, encoding desired behaviors as rewards is a challenging task leading to its separate field of study, namely reward shaping.
            Even using the same rewards, trained agents may have a high variance in behaviors learned.
            They may even depend on the seeds with which the environment is initialized.
            Due to issues with sample-complexity and the unsafety of the exploration process in Deep-RL algorithms, a simulation is often used for training agents.
            Using simulations affects robustness via what is known as the sim-to-real gap, namely the difference in dynamics between the defined simulation and reality.
            Due to such discrepancies, agents trained in simulation may fail to transfer to reality.
            Methods for tackling these issues include regularization\cite{caps}, reward shaping techniques \cite{reward_shaping, real}, and constrained MDPs \cite{cmdp}.
            One advantage of model-based Deep-RL where models are learned, is that a vast array of pre-existing machine learning literature around the problem of generating models which closely approximate reality exists.
            Since NNs are data-hungry, common techniques include pretraining a model on simulation data, and then fine-tuning it to data observed in reality.
            Domain adapting models has its own complications, for example, this work \cite{trash} represents a modern challenge where trash on a conveyor must be segmented for recycling purposes.
            Even though large amounts of data in simulation are available and an acceptable number of data points in reality exist, current techniques still fail to achieve desirable performance.
            Adding safety to RL methods, whether by proving that policies will never explore unsafe regions or showing that optimal controllers are Lyapunov stable, is a recent research focus in its infancy.
            The following survey \cite{safe_rl} discusses important techniques within the emerging field of safe Reinforcement Learning.

    \subsection{Classical Control Theory Augmentations}\label{sub:augmented_ct}
        Methods from classical control theory can be augmented with machine learning methods while allowing the use of a wide range of tools developed, from differential equation integrators \cite{Butcher_2008}, to automated proof search techniques \cite{Mitsch_Platzer_2016}.
        Some works improve classic system identification by adding a NN which can learn to model various nonlinear features but constrain the NN enough that guarantees are retained.
        An example of such work is explored in Section~\ref{sub:neural_lander}.
        Other works improve the stability of preexisting controller architectures by representing the stabilizing function as an NN.
        Works showing how this is achievable while retaining guarantees is explored in Section~\ref{sub:neural_lyapunov}.
        Some methods use machine learning techniques to find approximating dynamical models represented as differential equations.
        A method in this area that is garnering recent success is discussed in Section~\ref{sub:SINDY}

        \subsubsection{Neural Lander}\label{sub:neural_lander}
            The Neural Lander work of \citep{neurallander} augments classical quadrotor dynamical models with a deep-NN to improve the accuracy in controlling and landing a quadrotor at precise points.
            Their goal is to do so while retaining classical control theory guarantees, namely the \textbf{global exponential stability} (see [def.a] in Section 2.4.1) of the system.
            In this case, the difficulty in achieving high control accuracy lies in the \note{complexity in modeling \textit{ground-effects}} affecting the system's dynamics.
            Their evalutations in Section VI C, show that their NN-augmented dynamical model provides significant improvements in prediction accuracy of near ground trajectories against existing ground effect models.
            However, they note in Section VI E, that without constraining this learned model, synthesized controllers produced unexpected outputs sometimes leading to crashes.
            This motivates their method's requirement in proving global exponential stability as a safety requirement for the controller.
            Finally, the authors show great improvements in trajectory tracking and landing near the ground (Fig. 3 in \cite{neurallander}) most notably touting 10x improvements in landing accuracy (errors of 0.072m for the baseline against 0.007m for their method).
            We will apply their method to the simplified problem defined in Section.~\ref{sub:oven} of this document, showcasing how their different components interact.

            Assuming that our idealized model of the oven's temperature does not capture all of the dynamics present, following Section II in \cite{neurallander} we extend our ODE with the term $f_a$ representing external disturbances in temperature.
            \note{The authors specified $f_a$ as representing the disturbances generated by ground effects.
            The paper compares the trained NN with a defined ODE \cite{ground_effects} modeling these ground effects, and tout large improvements with respect to prediction accuracy in section V.C}
            \note{For our example, the disturbances} can be the type of food in the oven, or if \note{someone} open the oven door.
            Our new dynamics are as follows:
            $$ \dot{temp(t)} = -k(temp(t) - u(temp_\text{on} - temp_\text{off}) + temp_\text{off}) + f_a$$

            We then train the NN approximating $f_a$ represented by $\hat{f}_a(temp(t), u, \theta)$ such that it predicts the amount the system will be disturbed by within some $\Delta t$ ($\theta$ being the paramters of the NN) , following what was done in Section II and III of \cite{neurallander}.
            In order to gather the required training data we collect observations along trajectories the oven's temperature takes while an operator controls $u$ (similar to Fig. 1(b) of \cite{neurallander}) and then we subtract the values that our ode predicts to produce a dataset made up of $temp(t)$ and $u$ measurements, and the resultant measured $f_a$.
            If we only optimize our NN based on this dataset without constraining it, then $\hat{f}_a$ can conceivable take any values, then the Lyapunov function proving that our controller $c$ is globally exponentially stable no longer holds.
            In order to regain stability guarantees we will (following Section III. \cite{neurallander}) constrain the NN's Lipschitz constant by performing Spectral Normalization.
            Our optimization objective is now:
            \begin{align*}
              \underset{\theta}{\text{minimize}}\hspace{1cm} & \sum^T_{t=1} \frac{1}{T}||\hat{f}(temp(t), \theta) - f_a||_2\\
              \text{subject to}\hspace{1cm} & ||f||_{\text{Lip}} \leq \gamma
            \end{align*}

            As in Section IV B, We must now update our controller $c$ counteracting the NN's predicted disturbance.
            \begin{align*}
                -k(temp(t) - u(temp_\text{on} - temp_\text{off}) &= -\hat{f}_a(temp(t), u, \theta) \\
                u_t &= \frac{temp(t) - \hat{f}_a(temp(t), u_{t-1}, \theta)/k}{(temp_\text{on} - temp_\text{off})}
              \end{align*}

            Since the solution of the control parameter $u_t$ depends on previous action $u_{t-1}$, the authors solve for $u$ using a fixed-point iteration method, they make use of the Lipschitz constant of the neural network to prove that this method converges to a unique value of $u$ which finds the optimal action (Section V A).

            Now, in order to prove the full stability properties of the new controller, the authors make a couple of assumptions, the first assumption describes a limitation in how much the desired trajectory is allowed to vary.
            In our oven example, we want our system to reach a fixed $temp_\text{desired}$ therefore this assumption trivially holds.
            The second assumption is on the discrete-time control rate of different system components, in our oven example this would be the rate at which we get to observe $temp(t)$ and take action $u$ accordingly.
            The third assumption is the strongest one, that is, $|f_a - \hat{f}_a|$ is upper bounded.
            This means that the authors expect the prediction error between the NN and reality to never exceed a finite value $\epsilon$.
            This assumption is a strong one because this means the NN must generalize to points unseen in the training set, and given the prevalence of adversarial examples, further study is required to show whether this is a fair assumption with respect to a Lipschitz-constrained NN.
            The authors then proceed with proving that with the defined assumptions, the controller is exponentially stabilizing towards an $O(\epsilon)$ sized ball around the set-point.
            They do so using \textbf{control Lyapunov functions} to prove this stability property (see [def.b] in Section 2.4.1, where the oven's Lyapunov conditions are proven)

            Then the authors proceed to test their method (Section VI), in the oven case, this involves comparing the pure controller $c$ against $c$ with the disturbance rejection algorithm.
            This requires our oven to invoke the NN multiple times (due to the iterative solver) in order to converge to an optimal value for $u$, thus needing an embedded system with large enough resources on board to achieve timely execution in satisfying the second assumption.
            By calculating the amount of error aggregated while the oven reaches our desired temperature, we can get measurements to be used as experimental results.
            For the quadrotor landing case, the authors perform multiple experiments, including: Trajectory tracking near the ground, Landing at certain points, and trajectory tracking which partially passes over a table.
            All their results show significant improvement in accurately following the required paths.

        \subsubsection{Neural Lyapunov Functions and dReal}\label{sub:neural_lyapunov}
            The work by \citeauthor{neural_lyapunov} presents an algorithm for finding NNs which define control Lyapunov functions.
            In doing so, the authors make use of the powerful function approximation capabilities of deep-NNs to characterize a controller's region of attraction.
            The authors focus on control Lyapunov functions for their capability in proving stability in even non-linear dynamical systems.
            
            We've proven that our example oven with controller $c$ is already stable over all of the state space (global stability) in [def.b] of Section 2.4.1 via defining a corresponding Lyapunov function.
            However, if our example system was a Segway, then there exists points in the state space from which reaching our goal is impossible for any control input $u$.
            Thus, we can consider the system stable (eventually reaching our goal state) under a subset of the possible state-space.
            This is termed the region of attraction, this notion is made precise in definition 7 of \cite{neural_lyapunov}.
            
            Depending on the choice of Lyapunov functions, we may underrepresent the region of attraction since the Lyapunov conditions may only hold under a strict subset of the true region of attraction.
            When the dynamical system or controller are complex or even changing, relying on an expert to find Lyapunov functions is impractical.
            This is the reason methods for automatically finding Lyapunov functions have been studied, but previous techniques constrain its mathematical structure (ex. a polynomial in the state), and thus limit the expressiveness in representing this region of attraction.
            Thus, the authors make use of deep learning techniques for finding Lyapunov functions allowing for greater freedom in our choice of controller as well (which can also now be a NN) while maintaining full stability guarantees with respect to the dynamical model.
            To that end, the authors define the notion of \textbf{Lyapunov risk} in Section 3.1.
            It measures the amount the Lyapunov conditions ([def.b] of Section 2.4.1 of this document) are violated.
            We relate the discrete-time Lyapunov conditions defined in [def.b] to their Lyapunov risk equivalent as follows:
            \begin{align*}
                V(temp_\text{desired}) = 0 &\longrightarrow V_\theta(temp_\text{desired})^2\\
                \forall_{x\in\mathbb{R/} \{temp_\text{desired}\}}V(x) > 0 &\longrightarrow \max(-V_\theta(x),\epsilon)\\
                \forall_{x\in\mathbb{R}}, V(xe^{-k(\Delta t)}+temp_\text{desired}) \leq V(x) &\longrightarrow \max(V_\theta(x) - V_\theta(xe^{-k(\Delta t)}+temp_\text{desired}), 0)
            \end{align*}
            
            We can then define a loss function representing the total Lyapunov risk computed over a sampled dataset, in the case of the oven, this would be samples $x_i \sim U(50, 400)$ (a uniform distribution over the operating temperatures of the oven).            
            $$
            \underset{\theta}{\text{minimize}}\hspace{0.8cm}\frac{1}{N} \sum_{i=1}^N\left(\max \left(\epsilon, -V_\theta\left(x_i\right)\right)+\max(0, V_\theta(xe^{-k(\Delta t)}+temp_\text{desired}) - V_\theta(x))\right)+V_\theta^2(temp_\text{desired})
            $$
            Notice that this loss reaches 0 for all possible samples if the Lyapunov conditions are met.
            However, fully optimizing this loss function does not guarantee that the NN satisfies the properties over the whole state-space which may be continuous.
            We therefore define a falsification constraint similar to example 1 in \cite{neural_lyapunov}:
            $$\Phi_\epsilon(x): (V(temp_\text{desired}) = 0)\ \land\ (V(x) > \epsilon)\ \land\ \left(V(xe^{-k(\Delta t)}+temp_\text{desired}) - V(x)\leq 0\right)$$
            
            The authors therefore turn to SMT solvers for finding falsifying counter examples (an $x$ such that $\lnot\Phi_\epsilon(x)$) and adding them to the dataset.
            But, the problem of checking the Lyapunov conditions when non-linear operators are used is in full generality undecidable.
            This is why the authors use the work of \citeauthor{dreal} which provides decision procedures for first-order nonlinear theories over the reals and a corresponding SMT solver called dReal.
            
            dReal makes the problem of checking our previously mentioned conditions decidable via allowing false negatives in a $\delta$-sized region around our conditions, turning the problem NP-complete.
            For example checking the condition $x = 0$ becomes checking $|x| < \delta$, for some configurable $\delta$ parameter trading off computational time with accuracy.
            dReal supports real-valued linear and exponential functions, allowing the Neural Lyapunov work to use NNs whose architectures are composed of sigmoid activations and linear transformations.
            dReal being $\delta$-complete (definition 6 in \cite{neural_lyapunov}) means that when the algorithm accepts the statement, all the conditions are met.
            When dReal rejects the statement, it provides falsifying examples which we would add to our data-set as well as some false negatives.
            However, this would only add extra points to the training set which does not affect the correctness of the authors' proposed algorithm (algorithm 1 in \cite{neural_lyapunov}).

            When dReal accepts the formula $\Phi_\epsilon$ as fully satisfied, the algorithm terminates guaranteeing the stability of the associated controller.
            The authors then showcase uses of their method by allowing controllers to be optimized with respect to the learned Lyapunov functions.
            They then evaluate the region of attraction for these controllers in their experimental section, showing in some cases a $600\%$ improvement on the size of the region of attraction over the guarantees provided by commonly used LQR controllers.
            
            While they show that these techniques are applicable in some non-trivial systems, the problem being NP-complete in the number of neural network parameters limits this algorithm's tractability.
            Other work \cite{Dai_Landry_Yang_Pavone_Tedrake_2021} applying this method, has shown that finding a controller and corresponding Lyapunov function in a similar manner takes three days of computation on a powerful machine (Intel Xeon CPU).
            However, the importance of the method lies in enabling the capability of learning and guaranteeing the correctness of otherwise hard to interpret NN controllers, also forcing the lack of adversarial examples.
        \subsubsection{SINDY}\label{sub:SINDY}
            \textbf{Sparse identification of nonlinear dynamical systems} (SINDY) \cite{SINDY} is a method within the intuitive physics subfield.
            It tries to find a system's governing equation from data.
            The method defines and searches a space of differential equations defined by polynomials of the system's state vector while using candidate nonlinear operators.
            Similar to how kernel tricks are employed to capture nonlinear terms in polynomial regression.
            An L1 regularizer is used to bias the learned equations towards solutions with lesser terms, encouraging sparsity.
            The author hypothesizes that most physical systems have governing equations with primary components following simple laws.
            When this method finds good approximating models, they tend to generalize well to unseen data and are inherently explainable.
            An advantage of this method is that it unlocks the use of extensively researched tools built for control relying on differential equations.
        	However, the method also has significant limitations, such as allowing only separable nonlinear functions.
        	It disallows the use of multiple nonlinear operators whose first few terms in their Taylor series expansions are similar.
        	It also fails to scale to large state spaces such as images.
        	Later introduced techniques such as applying SINDY in the latent space of NN encoders \cite{latent_SINDY} have been explored to address this limitation.


        


\section{Embedded System Design Considerations}\label{sec:embedded_design}
    Robotic systems have a wide range of capabilities and resource limitations, from kilobots \cite{kilobots} to the Saturn V rocket \cite{saturn_v}.
    Common software design goals include having low system unpredictability, low system latency, low power usage, and high performance where needed.
    When designing such systems, controller safety guarantees rely on assumptions of timeliness and predictability of the embedded systems on which they are expected to be deployed.
    In Section~\ref{sub:unpredictability} we focus on sources of uncertainty in idealized models in reference to what occurs on real systems.
    In Section~\ref{sub:hard_soft_design} we discuss hardware and software designs to maintain controller guarantees.
    \subsection{Sources of Unpredictability}\label{sub:unpredictability}
        When modeling stochastic systems, there are two central notions of uncertainty aleatoric and epistemic.
        (i) Aleatoric uncertainty comes from the system's unpredictability with respect to what is observable from the robots' sensors.
        This type of uncertainty includes uncapturable system dynamics due to missing measurements, chaos where the system's trajectory may be unpredictable, and other vital factors like control-rate issues intruding with the expected timing of sensor readings or controller actuation.
        (ii) Epistemic uncertainty where the model itself causes errors.
    	For example, a differential equation may simplify or ignore factors such as air resistance.
    \subsection{Hardware and Software Design}\label{sub:hard_soft_design}
        The required control rate and the number of executed operations per control loop vary depending on the method used and the dynamical system's behavior.
        Also, any lag or jitter in executing control tasks on such systems increases the approximation errors of the dynamic models in question.
        Controller design may then be overly conservative, taking into account the increased aleatoric uncertainty due to the added model approximation errors, otherwise risking violating defined safety criteria.
        Therefore, enough computational resources must be reserved for control when designing hardware for such systems.
        In simplistic systems (e.g., single-core systems), it may be enough to provide a computational budget based on the worst-case execution time of specified control tasks.
        Federated architectures~\cite{federated} allow this type of reasoning to scale to systems with many predefined tasks, isolating them by separating hardware modules.

        Motivated by reducing Size, Weight, Power, and Cost (SWaP-C), modern embedded systems have multi-core architectures with complex memory layouts, relying on the operating system to manage resources and schedule defined tasks.
        However, due to the previously mentioned hardware limitations, tasks must share resources when executing, forcing in many cases the loss of compositionality with respect to individual task resource usage requirements.
        For example task-interference may cause deadline-misses which forces the system to lose some timing guarantees.
        The identification of such issues is a topic of heavy research within the embedded systems community and is surveyed in the following work~\cite{embedded_interference}.
        
        There has been much real-time operating system research in obtaining timing guarantees on these systems, theoretically mitigating the aforementioned problems.
        These considerations include handling resource contention, choosing scheduling algorithms, task-to-resource mapping techniques, and worst-case execution time estimation techniques.
        Such works are described in the following survey~\cite{embedded_timing_survey}.
        Then there are works which implement these techniques on actual systems and evaluate how well these theoretical guarantees hold.
        Such works include the Quest Operating System~\cite{quest} which is a real-time OS providing timing guarantees with respect to cpu-intensive tasks, Quest-V~\cite{quest_v}, Jailhouse~\cite{jailhouse}, and BAO~\cite{bao} which are partitioning hypervisors allowing for tasks to use virtually isolated resources. Example works implementing virtual memory isolation are Palloc~\cite{palloc} and Memguard~\cite{memguard}. The details of these techniques are outside the scope of this survey paper.

        Separately, the work done in \textbf{smARTflight} \cite{smartflight} shows that even in simple cases, making scheduler design choices along with the respective control algorithm allows for gains to be made in terms of performance and energy efficiency.
        The authors observe significant jitter in the control rate of a standard firmware used for flight control, namely CleanFlight \cite{cleanflight}.
        As discussed in detail in Section 4.2 of the paper, the issue of unpredictability and non-determinism within the defined control tasks defined leads to negative effects in quadrotor response-time and energy efficiency, stemming from the lack of timing guarantees in the standard CleanFlight best-effort scheduler.
        The authors then modify the firmware by including a preemptive criticality-aware rate-adaptive scheduler.
        They then make use of these capabilities and define two PID control tasks with varying rates, one actuating at 250Hz considered the energy efficient mode, and another actuating at 1000 Hz as the high-performance mode.
        This allows for the controller to trade responsiveness and energy efficiency based on the attitudes the quadrotor must reach.
        The authors then evaluate these results showing a 60\% improvement in response time in certain scenarios.

\section{Conclusion}
In achieving safety, it's important for the behavior of controlled systems to be well understood.
Safety benefits from more predictable systems (e.g. systems with timing guarantees), and from well understood controller specification (e.g. control Lyapunov functions).
Thus, tying the understanding of classical methods with the new capabilities offered by recent machine learning techniques allows for a compromise worth exploring further.
Existing methods combining specification techniques, learning, and control, though, are limited by the practicality of the algorithms in use, and the blackbox nature of common machine learning algorithms makes the trade-off between understanding and representative power difficult to balance.
The NeuralLander work~\cite{neurallander} is limited by having to collect a large dataset from the real world before being able to use the controller, and the heavy assumption on NN generalisability, in limiting the capability of the NN, they are able to retain human interpretability, along with formal stability guarantees.
Extending this method so that we train the NN online and update the controller while retaining safety poses an interesting research direction.
The NeuralLyapunov method is limited by the computational complexity required for finding NN certificates, even on modern machines, quadrotor control hits the limit of practicality for this method.
It is possible that the SMT solving step, which makes sure there are no counter-examples, does not affect the probability of failure in real systems, given that there is always a gap between the behavior of actual systems and the models of those systems.
There is a method which explores the notion of an Almost Lyapunov function \cite{almost_lyapunov}, which shows that stability is guaranteed even when the Lyapunov function isn't fully satisfied everywhere, but it is limited in having the state space be euclidean for their complex proof to apply.
Also worth exploring, is the addition of extra gradient signals for speeding up the stochastic-descent search for Lyapunov certificates.
One could even reuse RL techniques so that users can specify rewards which guide controller/Lyapunov function search.

Recent work on differentiable specifications allows for defining logical rules for controllers to follow (defining precisely what safety means in certain contexts).
But further work is required in this area as Differentiable Signal Temporal Logics, for example, do not capture the notion of trade-offs, meaning how important one criteria is against the others.
Finally, there aren't enough works which connect safety information across the whole stack, from embedded system's capabilities to the NN architecture. For example the smARTflight work can be extended via introducing NN controllers which are capable of using information from the scheduler, such as the predicted and previous $\Delta t$ between invocations.
Making the model and NN controller "aware" of the mode they are running in, retaining safety while maximizing performance or efficiency depending on the mode.
\color{black}
\bibliography{refs}

\end{document}